%% file: main.tex
\documentclass{article}
\usepackage{ijcai23}

\usepackage{times}
\usepackage{soul}
\usepackage{url}
\usepackage[hidelinks]{hyperref}
\usepackage[utf8]{inputenc}
\usepackage[small]{caption}
\usepackage{graphicx}
\usepackage{amsmath}
\usepackage{amsthm}
\usepackage{booktabs}
\usepackage{algorithm}
\usepackage{algorithmic}
\usepackage[switch]{lineno}

\usepackage{amsfonts}       
\usepackage{amsmath}
\usepackage{todonotes}
\usepackage{subcaption}
\usepackage{bm}
\usepackage{enumitem}   

\usepackage{calc}
\newlength\myheight
\newlength\mydepth
\settototalheight\myheight{Xygp}
\newcommand*\inlinegraphics[1]{%
  \settototalheight\myheight{Xygp}%
  \settodepth\mydepth{Xygp}%
  \raisebox{-\mydepth}{\includegraphics[height=\myheight]{#1}}%
}

\usepackage{siunitx}

\input{macros.tex}

\newtheorem{example}{Example}

\newtheorem{definition}{Definition}

\title{Analyzing Intentional Behavior in Autonomous Agents under Uncertainty}

\author{
Filip Cano C\'ordoba$^1$
\and
Samuel Judson$^2$\and
Timos Antonopoulos$^{2}$\and
Katrine Bj{\o}rner$^{3}$\and\\
Nicholas Shoemaker$^2$\and
Scott J. Shapiro$^{2}$\and
Ruzica Piskac$^2$\And
Bettina K{\"o}nighofer$^1$
\affiliations
$^1$Graz University of Technology\\
$^2$Yale University\\
$^3$ New York University\\
\emails
\{filip.cano, bettina.koenighofer\}@iaik.tugraz.at,
\{samuel.judson, timos.antonopoulos, nick.shoemaker, scott.shapiro, ruzica.piskac\}@yale.edu, kbjorner@nyu.edu
}

\begin{document}

\maketitle

\begin{abstract}
Principled accountability for autonomous decision-making in uncertain environments requires distinguishing intentional outcomes from negligent designs from actual accidents. 
We propose analyzing the behavior of autonomous agents through a quantitative measure of the evidence of intentional behavior. 
We model an uncertain environment as a Markov Decision Process (MDP). For a given scenario, we rely on probabilistic model checking to compute the ability of the agent to influence reaching a certain event. We call this the \emph{scope of agency}.
We say that there is evidence of intentional behavior if the scope of agency is high and the decisions of the agent are close to being optimal for reaching the event.  Our method applies counterfactual reasoning to automatically generate relevant scenarios that can be analyzed to increase the confidence of our assessment.
 In a case study, we show how our method can distinguish between `intentional' and `accidental' traffic collisions. 
\end{abstract}

\section{Introduction}
\let\thefootnote\relax\footnotetext{
Find code and experimental details in the accompanying repository \url{https://github.com/filipcano/intentional-autonomous-agents}.
}
Artificial intelligence (AI)-based autonomous agents play a significant role in diverse facets of society, such as transportation, robotics, medical devices, manufacturing, and more. Ideally, engineers would verify their correctness before deploying them in the real world. 
However, for various theoretical and practical reasons, formal verification of software for autonomous agents is not often feasible. 
As a result, autonomous agents might not behave as planned initially and they might cause harm. 
As we cannot predict when harm will happen, we need to examine the software of the harming autonomous agent \emph{ex post} -- after the harm -- to assess questions of accountability. 
While the liability scheme for autonomous agents has yet to be developed, it is plausible to assume that manufacturers of autonomous agents that intentionally harm should be held to a higher standard of accountability than ones that create agents that harm negligently or purely accidentally. 
Therefore, defining and understanding intention is of paramount significance for establishing accountability.  
In this paper, we propose a new way of determining whether an autonomous agent has, in fact, acted in a way consistent with the intention to harm. 

Historically, symbolic AI produced a large body of work to formally specify and design autonomous agents that were `rational'. Such agents would explicitly derive decisions based on their beliefs, desires, and intentions (BDI)~\cite{bratman1987intention,rao1995bdi}. Determining whether an autonomous agent has acted with the intention to harm is easy in the case of BDI agents. 
One just needs to read off the intentions from where they are written in the code. 
The statistical nature of modern machine-learning-based agents leaves the interpretation of their decision-making in probabilistic settings a far greater challenge,
since intentions are not explicitly present in such models.

The traditional view of intention establishes a connection to planning through either cognitive or computational reasoning. Intention is a nuanced legal and philosophical term of art. Here, we use it in the restricted sense of the `state of the world' the agent plans towards.
Whether human or machine, a rational agent with bounded resources must necessarily plan towards a goal to succeed in achieving it~\cite{bratman1987intention,cohen1990intention}. Modern machine-learned agents plan implicitly through techniques like reinforcement learning (RL)~\cite{sutton2018reinforcement}.

This paper considers an autonomous agent operating with other agents within an environment. During the agent's operation, a certain event happened. In the context of holding the agent accountable for such an event, we want to analyze whether the agent acted towards making that event happen. 

\paragraph{Problem statement.} We concretely model the interactions between the agent and its environment as a Markov Decision Process (MDP). 
The event under analysis is formalized as a set of states 
$S_\mathcal{I}$ in the MDP.
Our goal is to analyze whether the decision-making policy of the agent \emph{shows evidence of intentional behavior towards reaching $S_\mathcal{I}$}.

If we assume that the agent has perfect knowledge about the entire world as captured in an MDP, we could simply say: `There is evidence of intentional behavior towards reaching a state in $S_\mathcal{I}$, if the agent implements a policy that maximizes the probability of reaching $S_\mathcal{I}$'. However, for any agent acting within a complex environment, this assumption is implausible. For example, the current state information might not be precise due to imprecisions in sensor measurements, bounded resources in computing the policy, imprecisions due to abstractions,
partial observability, or usage of inaccurate models of other agents. 
Therefore, we need to consider a certain degree of uncertainty in our assessments.

\paragraph{Method for analyzing intentional behavior.}
In this paper, we propose a methodology to analyze 
whether there is evidence that an agent acted intentionally to reach a state in $S_\mathcal{I}$.
For a given scenario, we use probabilistic model checking to automatically compute the policies that maximize and minimize the probability to reach $S_\mathcal{I}$. 
We use these policies to compute the influence that the agent had to bring about $S_\mathcal{I}$.
We call this the \emph{scope of agency}.
We say that there is evidence of intentional behavior if the scope of agency is high and the decisions of the agent are close to optimal for reaching $S_\mathcal{I}$. 

To strengthen our evaluation, we make use of a
widespread technique in accountability analysis~\cite{wachter2017counterfactual}, which is analysing a diverse set of relevant \emph{counterfactual scenarios}, and aggregating the evaluation results. 

\paragraph{Main Contributions.} The main contributions of this paper are the following:
\begin{itemize}
    \item To the best of our knowledge, we present the first method that analyzes intentional behavior directly from the policies in MDPs.
    \item We give definitions for \emph{evidence of intentional behavior} in MDPs.
    \item We propose a method to analyze evidence of intentional behavior of agents in MDPs.
    Our method uses model checking to automatically relate the agent's policy to any other possible policy. 
    Furthermore, our method applies 
    \emph{counterfactual reasoning} to increase the reliability of the assessment. 
    \item We provide a case study in which we analyze potential intentional behavior in the same scenario for different implementations of driving agents.
\end{itemize}

\section{Preliminaries}\label{sec:prelims}
\subsubsection*{Markov Decision Processes}

A Markov Decision Process (MDP) is a tuple $\mathcal M = (\mathcal S, \mathcal A, \mathcal P)$, where 
	 $\mathcal S$ is the set of \emph{states}, 
	 $\mathcal A$ is the set of \emph{actions} and 
	$\mathcal P:\mathcal S \times \mathcal A \times \mathcal S \to [0,1]$ is the \emph{
 transition function}.
A state represents `one way the world
can exist',
so any information available to the agent for 
deciding what to do is included in the state of the MDP.
The set $\mathcal A$ contains every possible action that can be taken
by the agent. 
The function $\mathcal P$ represents the transition to a new environment state that is produced as the result of the agent executing an action in a particular state. 

A \emph{trace} is a finite or infinite sequence of states 
$\tau~=~(s_1,~s_2,~\dots)$.
A trace $\tau$ is \emph{valid}
if 
for each $i$, there exists at least one $a_i\in \mathcal A$
such that $\mathcal P(s_i,a_i,s_{i+1}) > 0$.

The agent is modeled by a memoryless and deterministic \emph{policy} 
$\policy\colon \mathcal S\to \mathcal A$ over $\mathcal M$ that assigns an action to each state.
In Section~\ref{sec:discuss} we discuss how our method can be extended to consider strategies with non-determinism and memory.

\subsubsection*{Probabilistic Model Checking}
Using probabilistic model checking~\cite{clarke2018handbook}, 
we can compute the exact probability $\mathcal P_{\pi}(\varphi,s)$ of $\pi$ satisfying a property $\varphi$ for each state $s$ of the MDP~\cite{kwiatkowska2011prism,hensel2022probabilistic}. This property $\varphi$ will typically be defined in a probabilistic variant of a modal temporal logic, like probabilistic linear temporal logic (PLTL)~\cite{pnueli1977temporal}.

Let $\Pi\subseteq \{\pi\colon\mathcal S\to\mathcal A\}$ be a set of policies. 
We denote the maximum probability of 
satisfying $\varphi$ restricted to a policy in $\Pi$
as 
$\mathcal P_{\max|\Pi}(\varphi,s) = \max_{\pi\in\Pi}\mathcal P_\pi(\varphi, s)$.
Similarly, we denote the minimum probability as 
$\mathcal P_{\min|\Pi}(\varphi,s)$.
In this paper $\varphi := \texttt{Reach}(S)$ encodes the property of \emph{reaching} any state $s$ in a set of states $S \subseteq \mathcal{S}$. 

\section{Definition of Intentional Behavior in MDPs}\label{sec:intention}

In this paper, we assume that we have given a scenario where a certain event happened, like the agent visited a certain location or the agent had a collision with another agent.
Our goal is to analyze whether there is evidence the agent intentionally acted towards reaching this event. 

In this section, we give the definitions for \emph{evidence of intentional behavior} of policies in the presence of uncertainty.
We use an MDP $\mathcal{M} = (\mathcal{S},\mathcal{A},\mathcal{P})$
to model the interaction of the agent and the environment.
In the following sections, we will then propose and implement a method to analyze intentional behavior according to the definitions of this section.

\subsection{Intentions of Agents with Perfect Information}

According to~\cite{rao1995bdi},  an \emph{intention} of an agent is a set of states $S_\mathcal{I}\subseteq \mathcal S$ the agent committed to reach. Therefore, the agent acts towards reaching $S_\mathcal I$  to the best of its knowledge.

Let us assume that the agent has perfect knowledge about the environment. 
For a set of states $S_\mathcal{I}\subseteq \mathcal S$ to be an intention of an agent, the agent has to implement a policy  $\pi$ that maximizes the probability of reaching $S_\mathcal{I}$. 
Formally, if
$S_\mathcal{I}\subseteq \mathcal S$ is an \emph{intention} of an agent, then 
$\mathcal P_{\pi}(\texttt{Reach}(S_\mathcal{I}),s) =  
\mathcal P_{\max|\Pi}(\texttt{Reach}(S_\mathcal{I}),s)$,
for any state $s\in \mathcal S$.

The policies considered to compute $\mathcal P_{\max}$
can be restricted to a set of policies $\Pi$, if there are policies that should be excluded for comparison. For example,
we may only be interested in policies for comparison that
satisfy certain properties like fairness or progress properties. 

\begin{definition}[Evidence of intentional and non-intentional behavior]
\label{def:intentionperfect}
An agent $\pi$ shows \emph{evidence of intentional behavior} in a state $s$ towards $S_\mathcal{I}$ if 
$\pi$ maximizes the probability of reaching $S_\mathcal{I}$, \ie,
$\mathcal P_{\pi}(\texttt{Reach}(S_\mathcal{I}),s) =  
\mathcal P_{\max|\Pi}(\texttt{Reach}(S_\mathcal{I}),s)$. Otherwise, we say that the agent has \emph{evidence of non-intentional behavior} in state $s$ towards $S_\mathcal{I}$.
\end{definition}

\subsection{Intentions of Agents Under Uncertainty}
The definition of intention given above assumes perfect knowledge about the environment and
the agent implementing a policy that is optimal for reaching $S_\mathcal{I}$.
However, if we want to fully analyze intentional behavior we have to take imprecision and uncertainties into account. 
Any agent operating in a complex environment needs to make abstractions about the environmental state and, most likely, only has partial observability. Furthermore, the agent has to make assumptions about the other agents that act within the environment, which may be incorrect. 
Therefore, we need to relax the definition of intention to take uncertainties into account. 

In order to analyze an agent $\pi$ under uncertainty, we first define the \emph{intention-quotient} $\rho_\pi(s)$ for a state $s$ which represents how close $\pi$ is to the policy optimal for reaching $S_\mathcal I$ from state $s\in \mathcal S$.

\begin{definition}[Intention-quotient at a state]
\label{def:intention-quotient}
For an agent $\pi$ at a state $s\in \mathcal S$, 
the \emph{intention-quotient} is defined as follows:
\begin{equation}
\label{eq:willingness}
\rho_{\pi}(s) = \dfrac{\mathcal P_{\pi}
(\texttt{Reach}(S_\mathcal{I}),s) - \mathcal P_{\min|\Pi}(\texttt{Reach}(S_\mathcal{I}),s)}
{\mathcal P_{\max|\Pi}(\texttt{Reach}(S_\mathcal{I}),s) -  
\mathcal P_{\min|\Pi}(\texttt{Reach}(S_\mathcal{I}),s)}.\nonumber
\end{equation}
\end{definition}

In contrast to the case of perfect information, 
the uncertainty in the agent's knowledge and resources 
implies uncertainty in the assessment of intentional behavior. 

\begin{definition}[Evidence of intentional and non-intentional behavior in states]
Given lower and upper thresholds
$0\leq\delta_\rho^L < \delta_\rho^U \leq 1$,
we say that there is \emph{evidence of
intentional behavior} towards the intention $S_\mathcal{I}$ in the state $s$, if
$\rho_{\pi}(s) \geq \delta_\rho^U$.
Analogously,
we say there is \emph{evidence of
non-intentional behavior} towards the intention $S_\mathcal{I}$ in the state $s$, if
$\rho_{\pi}(s) \leq \delta_\rho^L$.

In case that 
$\delta_\rho^L < \rho_{\pi}(s)  < \delta_\rho^U$,
we say that we have \emph{not enough evidence} for intentional behavior.
\end{definition}

By adjusting the thresholds $\delta_\rho^U$ and $\delta_\rho^L$, we can control how much discrepancy from the optimal policy under perfect information is allowed in order
for $\pi$ to be still considered as intentional or non-intentional behavior for $S_\mathcal I$.
In general, the higher the value of the intention-quotient $\rho_{\pi}(s)$, 
the more evidence the policy $\pi$ shows of intentionally trying to reach $S_\mathcal I$.
The lower the value of  $\rho_{\pi}(s)$, the more evidence the policy $\pi$ shows on acting 
without the intention to reach $S_\mathcal I$. 

An additional source of uncertainty is introduced by the scope of agency of a state.
In situations where the agent's actions have little effect on reaching $S_\mathcal{I}$,
there is not enough evidence to support a claim of intentional behavior. For this reason, we take the scope of agency into account for our definition of intentional behavior.

\begin{definition}[Scope of agency]
    The scope of agency $\sigma(s)$ at a state $s$ for intention $S_{\mathcal I}$
    is defined as the gap between the best and the worst policy in terms of reaching $S_\mathcal I$.
    Formally, it is given by
    \begin{equation}
    \sigma(s) =  \mathcal P_{\max|\Pi}(\texttt{Reach}(S_\mathcal{I}),s) -  \mathcal P_{\min|\Pi}(\texttt{Reach}(S_\mathcal{I}),s).\nonumber
    \end{equation}

    The \emph{scope of agency of a trace} $\tau$ is  given by 
    \[
        \sigma(\tau) = \frac1{|\tau|} \sum_{s'\in\tau}\sigma(s').  
    \]    
\end{definition}

If the scope of agency $\sigma(\tau)$ of a trace $\tau$ is very low,
any assessment about intentional behavior will be very weak.

The above definitions of intentional and non-intentional behavior 
apply to a single state in the MDP. 
In order to extend these definitions to traces in the MDP, 
we aggregate the intention-quotients of the individual states using the
scope of agency as the weighting factor.

\begin{definition}[Intention-quotient for traces]
    For an agent $\pi$ 
    operating along a trace $\tau$,
    the intention-quotient $\rho_\pi(S)$
    is given as 
    the weighted average
    \begin{equation}
        \rho_\pi(\tau) = \frac1{\sum_{s\in \tau}\sigma(s)} 
        \sum_{s\in \tau} \sigma(s) \rho_{\pi}(s).
        \nonumber
    \end{equation}
\end{definition}

\begin{definition}[Evidence of intentional and non-intentional 
behavior in traces]
    Given lower and upper thresholds
    $0\leq\delta_\rho^L < \delta_\rho^U \leq 1$,
    and an agency threshold $0<\delta_\sigma < 1$,
    we say that there is \emph{evidence of
    intentional behavior} towards reaching $S_\mathcal{I}$ along a trace $\tau$, if
    $\sigma(\tau) \geq \delta_\sigma$ and $\rho_{\pi}(\tau) \geq \delta_\rho^U$.

    We say that there is \emph{evidence of
    non-intentional behavior} towards reaching $S_\mathcal{I}$ 
    if
    $\sigma(\tau) \geq \delta_\sigma$ and $\rho_{\pi}(\tau) \leq \delta_\rho^L$.
    
    In case that $\delta_\rho^L < \rho_{\pi}(\tau) < \delta_\rho^U$ or $\sigma(\tau) < \delta_\sigma$, we say that we have \emph{not enough evidence} for intentional behavior.
\end{definition}

\section{Setting and Problem Statement}\label{sec:setting}
In this section, we describe the setting in which we want to analyze intentional behavior and give the problem statement.

\paragraph{Setting.}
We have a model of the environment in the form of an 
MDP $\mathcal M = (\mathcal S,\mathcal A,\mathcal P)$ 
that captures all relevant dynamics and possible 
interactions for an agent. 
We also have a concrete scenario to analyze in the form of a trace $\tau_{\textit{ref}} = (s_1,\dots, s_n)$. 
The trace $\tau_{\textit{ref}}$ is a sequence of visited states in $\mathcal{M}$ that leads to a state in $S_\mathcal I$,
\ie,  $s_n\in S_\mathcal I$.
The implementation of the agent is given in the form of a policy 
$\pi\colon \mathcal S\to\mathcal A$.
The underlying intentions of the agent are unknown.

\paragraph{Problem statement.}

Given this setting, 
we want to analyze whether there is evidence of the agent acting intentionally, 
with uncertainty thresholds 
$\delta_\rho^D$, $\delta_\rho^U$, and $\delta_\sigma$
for the intention-quotient and scope of agency, respectively.
Hence, we want to analyze whether there is 
\emph{evidence of intentional behavior} 
of the agent $\pi$
towards intention $S_\mathcal{I}$ in the scenario $\tau_{\textit{ref}}$.

\begin{example}\label{example}
    Let us consider a scenario in which
    an autonomous car collides with a pedestrian
    crossing the road.
    To analyze to which degree the car is accountable for the accident, we are interested in whether causing harm was the intention of the car.
    In such an example, $\mathcal{M}$ captures all relevant information necessary to analyze the accident, like positions and velocities of car and pedestrian, car dynamics, road conditions, etc.  
    The scenario $\tau_{\textit{ref}} = (s_1,\dots, s_n)$ is defined via the sequence of states prior to the collision.
    The set of states $S_\mathcal{I}$ represents collisions.
    We want to analyze whether the policy $\pi$ shows evidence of intentional behavior towards $S_\mathcal{I}$.
    To avoid unfair comparison with unrealistic policies, 
    we define a set of policies $\Pi$ that excludes 
    unreasonably slow-moving cars (e.g., cars that stop even though there is no other road user close by). 
\end{example}

\section{Methodology}\label{sec:method}

\begin{figure*}[t]
    \centering
    \includegraphics[width=\linewidth]{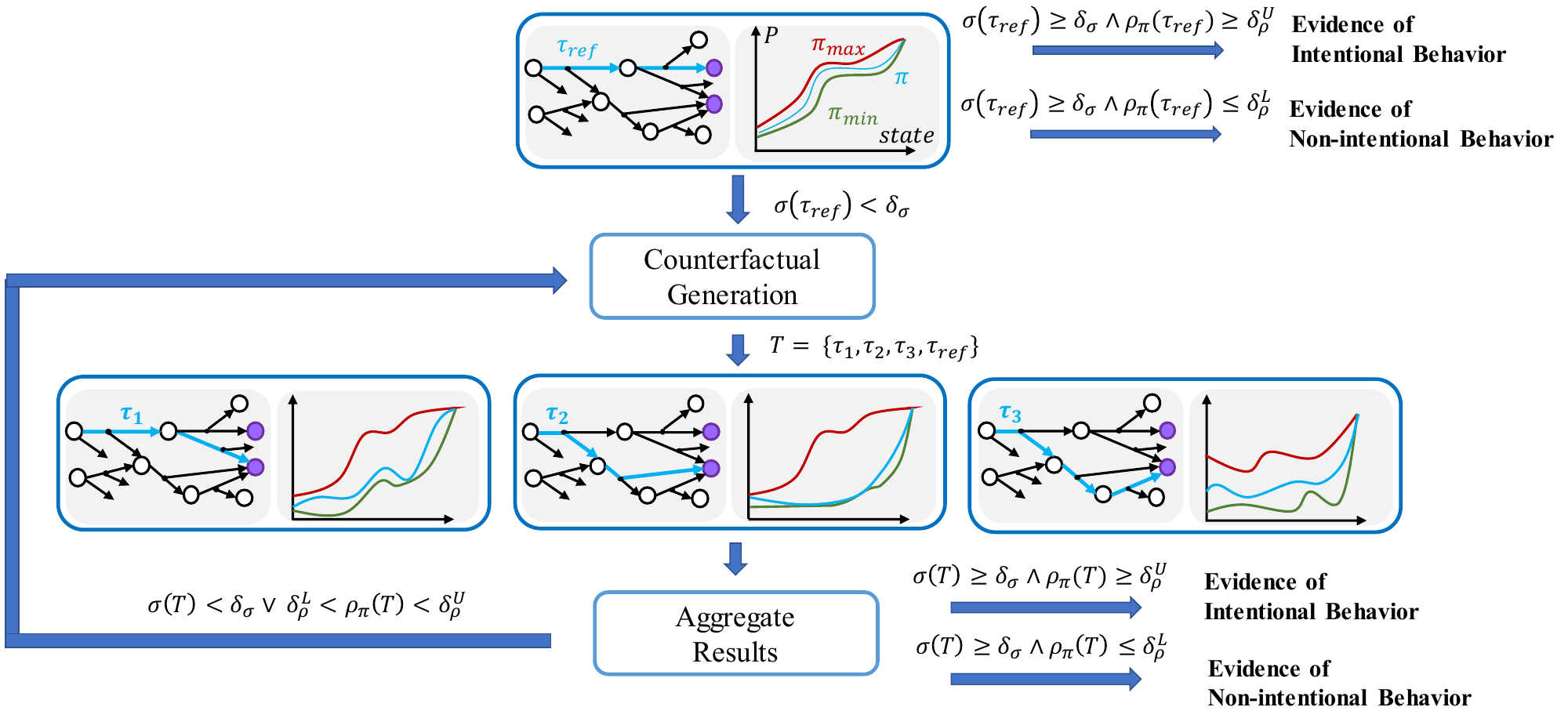}
    \caption{Overview of our approach to analyzing intentional behavior.}
    \label{fig:overview}
\end{figure*}

In this section, we propose a concrete methodology 
to analyze whether there is evidence an agent acted intentionally to reach $S_\mathcal I$.
Our method is illustrated in Figure~\ref{fig:overview}. 
As depicted in the figure,
we start the \emph{analysis of the given trace} $\tau_{\textit{ref}}$
by computing the intention-quotient $\rho_\pi(\tau_{\textit{ref}})$
and the scope of agency $\sigma(\tau_{\textit{ref}})$.
If $\sigma(\tau_{\textit{ref}}) \geq \delta_{\sigma}$, we can draw conclusions about intentional behavior:
\begin{itemize}
    \item If $\rho_\pi(\tau_{\textit{ref}}) \geq \delta_{\rho}^U$, then we conclude that there is evidence of \emph{intentional behavior} towards $S_\mathcal{I}$.
\item If $\rho_\pi(\tau_{\textit{ref}}) \leq \delta_{\rho}^L$, then we conclude that there is evidence of \emph{non-intentional behavior} towards $S_\mathcal{I}$.
\end{itemize}

In cases without enough agency, \ie, where $\sigma(\tau_{\textit{ref}}) < \delta_{\sigma}$, 
or where the intention-quotient falls between the lower and upper thresholds, \ie,
$\delta_{\rho}^L < \rho_\pi(\tau_{\textit{ref}}) < \delta_{\rho}^U$,
we say that we do have not enough evidence to reach a conclusion.
In such cases, we propose to \emph{generate} more evidence by analyzing \emph{counterfactual scenarios}.

A counterfactual scenario $\tau$ is a scenario close to $\tau_{\textit{ref}}$ according to some distance notion. 
Our method generates a set of counterfactual scenarios $T_{\textit{cf}}$ and 
computes whether there is evidence for intentional  
or non-intentional behavior
for each trace $\tau \in T = T_{\textit{cf}}\cup\{\tau_{\textit{ref}}\}$.
We fix beforehand the number of counterfactual scenarios to some parameter $N$.

As before, we draw conclusions about intentional behavior based on the \emph{aggregated results} of the scope of agency $\sigma(T)$ and intention-quotient $\rho_\pi(T)$. 
If 
$\sigma(T) < \delta_\sigma$ or $\delta^L_{\rho} < \rho_{\pi}(T) < \delta^U_{\rho}$,
there is still not enough evidence for intentional or non-intentional behavior,
with $\sigma(T)$ being the scope of agency 
averaged over all traces in $T$, 
and $\rho_{\pi}(T)$ being the average intention-quotient for the set of traces in $T$.

In such cases, 
our algorithm iterates back and extends the set $T_{\textit{cf}}$ by generating $N$ more counterfactual scenarios to be analyzed. 
The algorithm stops when enough evidence has been generated to draw a conclusion
or
when the number of generated counterfactual scenarios exceeds some user-defined limit.
In the following, we discuss the generation of counterfactual scenarios in detail.

\subsection{Counterfactual Generation}
In order to find enough evidence for our assessment of intentional behavior, we generate scenarios that are counterfactuals for $\tau_{\textit{ref}}$.
There are many ways to generate counterfactual traces.
We describe here three alternatives, ordered by decreasing the requirement 
of expert knowledge and involvement in the process.

\subsubsection*{Counterfactual Generation via a Human Expert}
Asking and analyzing counterfactual questions is a standard procedure in accountability processes~\cite{beebee2019counterfactual}. 
Usually, such counterfactual questions are proposed by a domain expert.
We transfer this concept
to analyzing intentional behavior on MDPs. 
The counterfactual questions posed by the expert are translated to counterfactual traces $T_{\textit{cf}}$ in the model  $\mathcal{M}$. 

\begin{example}
Recall Example~\ref{example}.
Some counterfactual questions posed by an expert in the traffic scenario
could be: 
(Q1) What if the car had driven slower? 
(Q2) What if the pedestrian had been visible earlier? 
(Q3) What if the road conditions were different?
Each of Q1-Q3 translates to a counterfactual trace, 
which we can analyze in our framework. 
\end{example}

The method of generating counterfactuals using a human expert imposes a heavy burden of work on the expert. 
Next, we propose two methods to automatically generate counterfactuals to mitigate the need for human effort.

\subsubsection*{Counterfactual Generation Using Factored MDP}
Since $\mathcal{M}$ models the interactions of the agent with its environment,
$\mathcal{M}$ is typically given in form of a \emph{factored} MDP. 
In factored MDPs, the state space of $\mathcal{M}$ is defined in terms of \emph{state variables} $\mathcal S = \mathcal{X}_1\times\cdots\times\mathcal X_m$.

In this approach for counterfactual generation, we assume domain knowledge about which variations of state variables generate interesting counterfactual scenarios. 
In particular, we assume to know which state variables are \emph{integral state variables} that we want to use in the analysis of intentional behavior, and which variables are
\emph{peripheral}.
To generate informative counterfactuals,
we are interested in changing the values of the integral state variables.

\begin{example}
    In Example~\ref{example}, integral state variables might represent the position and velocity of the car, the position of the pedestrian, the road condition, etc.,
    are integral variables. However, state variables that represent, for example, positions of other pedestrians located behind the car, are most likely labeled as peripheral by a human expert.   
    A counterfactual trace generated from changes in the pedestrians' positions that are not involved in the collision will give no new insights into the assessment of intentional behavior. On the contrary, changing the speed of the car might have a considerable effect on the collision probabilities and may provide an informative counterfactual scenario.
\end{example}

We automatically generate counterfactual traces by exploring variations of the integral variables. 
Let the state space be factored as $\mathcal{S} = \mathcal X_1\times\dots\times\mathcal X_m$, 
where variables $\mathcal X_1,\dots,\mathcal X_k$ are peripheral and $\mathcal X_{k+1},\dots,\mathcal \mathcal X_m$ are integral.
For any state $s=(x_1,\dots,x_m)$, 
we write its factorization 
into peripheral and integral variables 
as 
$s = (s^{\textit{per}}||s^{\textit{int}})$.
Let $s^{\textit{int}}_{\textit{ref}} = (x_{k+1},\dots,x_m)$ be the value of the integral variables 
at any state of $\tau_{\textit{ref}}$.
We define the set of counterfactual values as:
\begin{align}
    \mathrm{Cf_\epsilon}(s^{\textit{int}}_{\textit{ref}}) = &
    \{(y_{k+1}, \dots, y_{m}) \in \mathcal X_{k+1}\times\dots\times\mathcal X_{m}\::\: \nonumber\\
    \: & \qquad\forall i,\:
|x_i - y_i| < \epsilon_i 
\},\nonumber
\end{align}
where $\epsilon=(\epsilon_{k+1},\dots,\epsilon_{m})$
contains for each integral variable the range of variation that is still considered valid.
For a given trace $\tau_{\textit{ref}} = (s_1,\dots,s_n)$,
the counterfactual traces are
\begin{align}
    T_C(\tau_{\textit{ref}}) = & 
    \{(s'_1,\dots, s'_n)\::\:
    \exists s_{\textit{cf}}^{\textit{int}}\in \mathrm{Cf_\epsilon}(s^{\textit{int}}_{\textit{ref}}),\: \nonumber\\
    \: & 
    \forall i=1\dots n :
    s'_i = (s_i^{\textit{per}}||s_{\textit{cf}}^{\textit{int}}),\nonumber\\
    & (s'_1, \dots, s_n')\mbox{ is valid},\
   s'_n \in S_\mathcal I
    \}. \nonumber
\end{align}
Note that the search for counterfactual traces is limited to those integral variables
$\mathcal X_i$
for which $\epsilon_i >0$,
thus by setting some of the $\epsilon_i$ to zero, 
we can fix their value in the counterfactual generation process.

From $T_C$, we sample $N$ traces to be used for the counterfactual analysis. For the trace selection,
emphasis can be put on traces with higher scopes of agency.

\subsubsection*{Counterfactual Generation Using Distances on MDPs}

This method for generating counterfactual scenarios requires to have given a distance $d\colon \mathcal S\times \mathcal S \to \mathbb R_{\geq 0}$ defined over states in the MDP.  Given such a distance metric $d$ over the states, the set of counterfactual traces is given as
\begin{align}
    T_C(\tau_{\textit{ref}}) = & \{(s'_1,\dots, s'_n)\::\:
\forall i=1\dots n,\: d(s_i,s'_i) < \eta, \nonumber \\
   \: & (s'_1, \dots, s_n')\mbox{ is valid},\quad
   s_n \in S_\mathcal I\},\nonumber
\end{align}%
where $\eta>0$ is a distance that represents states being `close enough'
to be compared as counterfactuals.

In case there is no distance defined in the MDP, 
there are bisimulation distances that are well defined in any MDP~\cite{song2016measuring}.
They depend on the intrinsic structure of the MDP, 
defined mainly by similarities in terms of the transition function.
The main caveat of this approach is that distances are expensive to compute, and 
the explanation of why two states are assigned a given distance 
becomes more obscure to the user.

\section{Experimental Results}\label{sec:study}
\begin{figure}
    \centering
    \includegraphics[width=\linewidth]{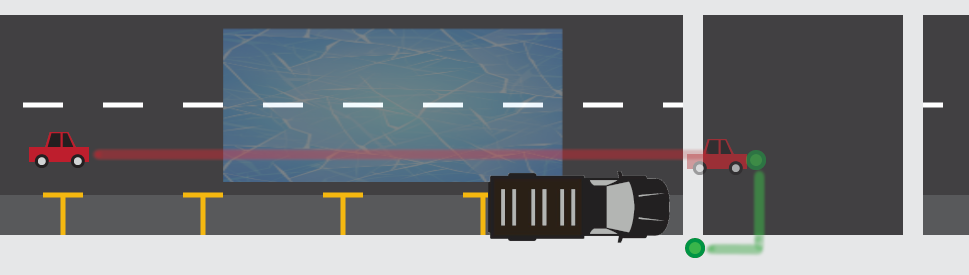}
    \caption{Case-study environment, with scenario $\tau_{\textit{ref}}$ highlighted.}
    \label{fig:casestudysetting}
\end{figure}

In this section, 
we showcase our method on a traffic-related scenario related to Examples 1-2, and that is illustrated in Figure~\ref{fig:casestudysetting}.
In this scenario, a car was driving on a road with a crosswalk. 
A pedestrian at the crosswalk decided to cross.
Close to the crosswalk, there was a parked truck that blocked the visibility
of the car. 
Furthermore, the cold weather conditions made the road slippery, 
so that braking was less effective than normal.
While crossing, the pedestrian was hit by the car.
We want to study the behavior of the car for signs of the hit being intentional.

All experiments were executed  
on an Intel Core i5 CPU with 16GB of RAM running Ubuntu 20.04.
We use \textsc{Tempest}~\cite{tempest}
as our model checking engine.

\subsection{Model of Environment}
The environment is modeled as an MDP 
$\mathcal M = (\mathcal S, \mathcal A, \mathcal P)$.
The set of states is a triple
$\mathcal S = \mathcal S^{\textit{car}} \times S^{\textit{ped}} \times S^{\textit{env}}$,
where $S^{\textit{car}}$ models the position and velocity of the car, 
$S^{\textit{ped}}$ models the position of the pedestrian,
and $S^{\textit{env}}$ models other properties 
that do not change during a scenario.
These properties include the slipperiness factor of the road and the existence of the truck blocking the car's view of the pedestrian.

The car's position is defined via the integers $x_c$ and $y_c$ with $0 \leq x_c \leq 60~\unit{\metre}$
and $3 \leq y_c \leq 13~\unit{\metre}$. The velocity of the car is in 
$\{0, 1, \dots, 5\}~\unit[per-mode=symbol]{\metre\per\second}$. 
 The position of the car is updated at each step,
assuming a uniform motion at the current velocity.
The car has the following set of actions $\mathcal{A}$:
hitting the brakes, pressing down on the accelerator, and coasting.
If the car is on a non-slippery part of the road, 
accelerating stochastically increases the velocity 
(by $1$ or $2~\unit[per-mode=symbol]{\metre\per\second}$), 
braking stochastically decreases the velocity  
(by $1$ or $2~\unit[per-mode=symbol]{\metre\per\second}$) and coasting maintains or decreases the velocity 
(by $1~\unit[per-mode=symbol]{\metre\per\second}$).
If the car is on a slippery part of the road, the 
consequences of the selected action on the velocity change, and include the possibility of no modification to the current velocity for both the actions of braking and accelerating.

The pedestrian's position is given via the integers $x_p$ and $y_p$ with
 $0 \leq x_p \leq 60~\unit{\metre}$ and $ 0 \leq y_p \leq 15~\unit{\metre}$. 
The pedestrian can move $1$m in any direction, or not move at all. 
The probabilities of moving in each direction are given by 
a stochastic model of the pedestrian, 
designed in such a way that the pedestrian favors crossing the street
through the crosswalk while avoiding being hit by the car.
The probabilities in the pedestrian's position update can be influenced by a hesitance factor, which captures how likely it is that the pedestrian puts themselves at a hitting distance from the car. 
The resulting MDP consists of about $120\mathrm{k}$ states and $400\mathrm{k}$ transitions.

\subsection{Analysis of a Trace}
\label{sec:analysis_trace}
In the described environment, we are given a scenario $\tau_{\textit{ref}}$
as illustrated in Figure~\ref{fig:casestudysetting},
and an agent $\pi\colon \mathcal S\to\mathcal A$.
As thresholds to evaluate evidence of intention, 
we use $\delta_\rho^L = 0.25$, $\delta_\rho^U = 0.75$ and 
$\delta_\sigma = 0.5$. We restrict the set of policies $\Pi$ to policies that do not stop the car if no pedestrian is within a range of $15$m of the car.
The collision states are given by 
$S_\mathcal{I} = \{s \in \mathcal{S} \::\: |x_p-x_c| \leq 5 \vee |y_p-y_c| \leq 5 \}$.
Given this setting, we analyze $\tau_{\textit{ref}}$ for evidence 
of intentional behavior towards reaching $S_\mathcal{I}$.
Therefore, we first compute the intention-quotient $\rho_\pi(\tau_{\textit{ref}})$
and the scope of agency $\sigma(\tau_{\textit{ref}})$.

\begin{figure}
    \centering
    \includegraphics[width=0.96\linewidth]{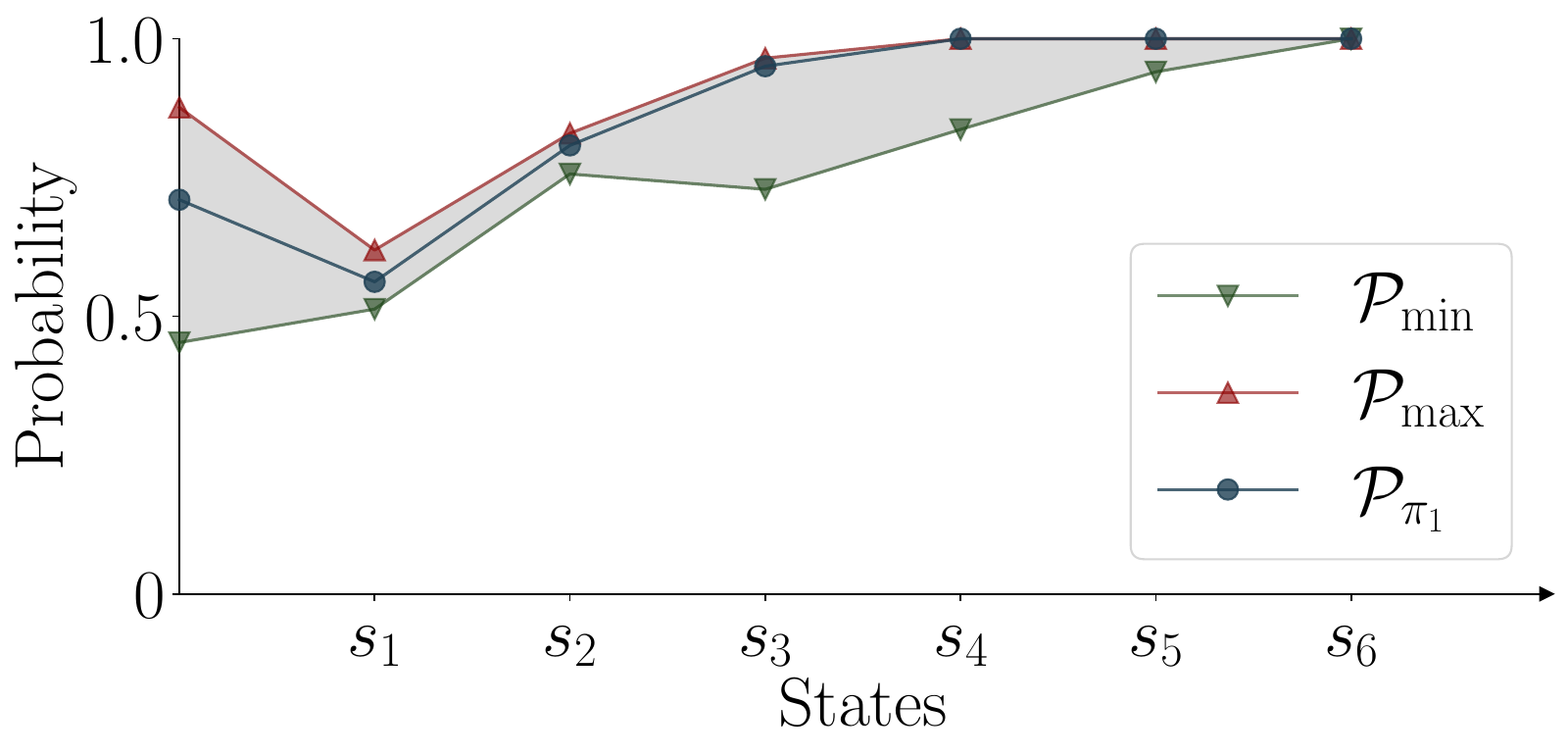}
    \caption{Intention-quotient and scope of agency of $\tau_{\textit{ref}}$.}
    \label{fig:tauref}
\end{figure}

\paragraph{Results of analysing $\bm{\tau_{\textit{ref}}}$.}
In Figure~\ref{fig:tauref}, we give the results of the model checking calls for reaching $S_\mathcal{I}$  for states in $\tau_{\textit{ref}}$.
The lower line (\inlinegraphics{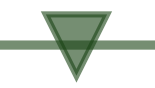})
represents $\mathcal P_{\min}$, 
the upper (\inlinegraphics{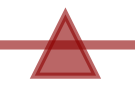}) represents $\mathcal P_{\max}$
and the line in the middle (\inlinegraphics{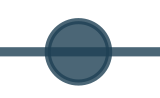}) represents $\mathcal P_\pi$ for every state in $\tau_{\textit{ref}}$.
The shaded area, 
between $\mathcal P_{\min}$ and $\mathcal P_{\max}$,
represents the scope of agency at each state. 
The figure shows the agent is close to the line of $\mathcal P_{\max}$,
but the scope of agency is very low,
with
$\rho_\pi(\tau_\textit{ref}) = 0.73$ and 
$\sigma(\tau_\textit{ref}) = 0.18$.
Since $\sigma(\tau_\textit{ref}) < \delta_{\sigma}$, our method concludes that there is not enough evidence for intentional behavior yet and moves on to the step of generating counterfactual scenarios.

\paragraph{Counterfactual analysis.}
We generate counterfactual scenarios by exploiting domain knowledge about integral variables of the MDP.
We change the following variables:
\begin{itemize}
    \item \emph{Slipperiness range}. The street is considered to be slippery between 
 the positions $sl_{\textit{init}}$ and $sl_{\textit{end}}$.
 \item \emph{Slipperiness factor}. The strength of the slippery effect is measured by the
 slippery factor $sl_{\textit{fact}}$,
 which is analogous to the inverse of the friction coefficient in classical dynamics.
 The effect of slipperiness is to 
 make the acceleration and brake less effective, 
 increasing the probability that both acceleration and brake 
 have no effect on the speed of the car.
 The larger the value of $sl_{\textit{fact}}$, 
 the more effect,
 with $sl_{\textit{fact}}=1$ being the minimum value, 
 where the road is considered to be `not slippery at all'.
 \item \emph{Hesitancy factor}. The pedestrian, in general, tends to cross the street 
 through the crosswalk. The hesitancy factor modifies the probabilistic model of the pedestrian, to make them more or less prone to put themselves at a hitting distance from the car. 
 A pedestrian with hesitancy factor $h_{\textit{fact}}=0$ is a completely cautious pedestrian.
On the contrary, a pedestrian with hesitancy factor
 $h_{\textit{fact}}=1$ completely disregards the state of the car.
 \item\emph{Visibility}. In the given scenario, there is a truck blocking the visibility of the car, corresponding to $vis=1$. 
 In case $vis=0$, the visibility block is eliminated.
\end{itemize}
The variables and the ranges considered for generating counterfactuals are summarized in Table~\ref{tab:ranges}.
\paragraph{Results of analyzing counterfactual scenarios.}
We build the counterfactuals in batches of $N=5$,
by sampling uniformly on the ranges described in Table~\ref{tab:ranges}.
We show the results in terms of intention-quotient and scope of agency in Table~\ref{tab:counterfactuals}. 
We report the averaged values and standard deviations over 5 runs. 
As we can see from the table, 
with 21 traces in $T$ we have $\rho_{\pi}(T) > \delta_{\rho}^U = 0.75$
and $\sigma_{\pi}(T) > \delta_{\sigma} = 0.5$. 
Thus, our method concludes that the agent under study does present evidence of intentional behavior to hit the pedestrian.

\begin{table}[t]
\centering
\small
\begin{tabular}{lccccc}
& $sl_{\textit{init}}$ & $sl_{\textit{end}}$ & $sl_{\textit{fact}}$ & $h_{\textit{fact}}$ & $vis$      \\
\midrule
Value $\tau_{\textit{ref}}$ & 20                   & 45                  & 2.5                  & 0.5                 & 1          \\ 
Range      & $[10, 30]$           & $[35,55]$           & $[1, 4]$             & $[0.1, 0.9]$        & $\{0, 1\}$\\
\end{tabular}
\caption{Ranges to use in counterfactual generation.}
\label{tab:ranges}
\end{table}

\begin{table}[t]
\small
\centering
\begin{tabular}{lrrrr}
$|T|$ & 6 & 11 & 16 & 21   \\
\midrule
$\rho_\pi(T)$    & 0.78 (0.03)  &  0.81 (0.02) &  0.83 (0.02)  &  0.84 (0.01) \\
$\sigma_\pi(T)$  &  0.33 (0.02)  &  0.44 (0.03)  &  0.48 (0.01) &  0.50 (0.01) \\
time (s)      &  53 (16)    &  147 (42)  &  227 (32)  & 318 (64) \\ 
\end{tabular}
\caption{Results of the counterfactual evaluation.}
\label{tab:counterfactuals}
\end{table}

\subsection{Comparative Analysis of Several Agents}
In this section, 
we illustrate how our method can be used to compare different agents 
in terms of intentional behavior.
We compare three different agents $\pi_1, \pi_2, \pi_3$ in the same scenario $\tau_{\textit{ref}}$. 
The agent $\pi_1$ corresponds to the policy $\pi$ in Section~\ref{sec:analysis_trace}.

\begin{figure}[t]
     \centering
     \begin{subfigure}[t]{0.26\textwidth}
         \centering
         \includegraphics[height=3.7cm]{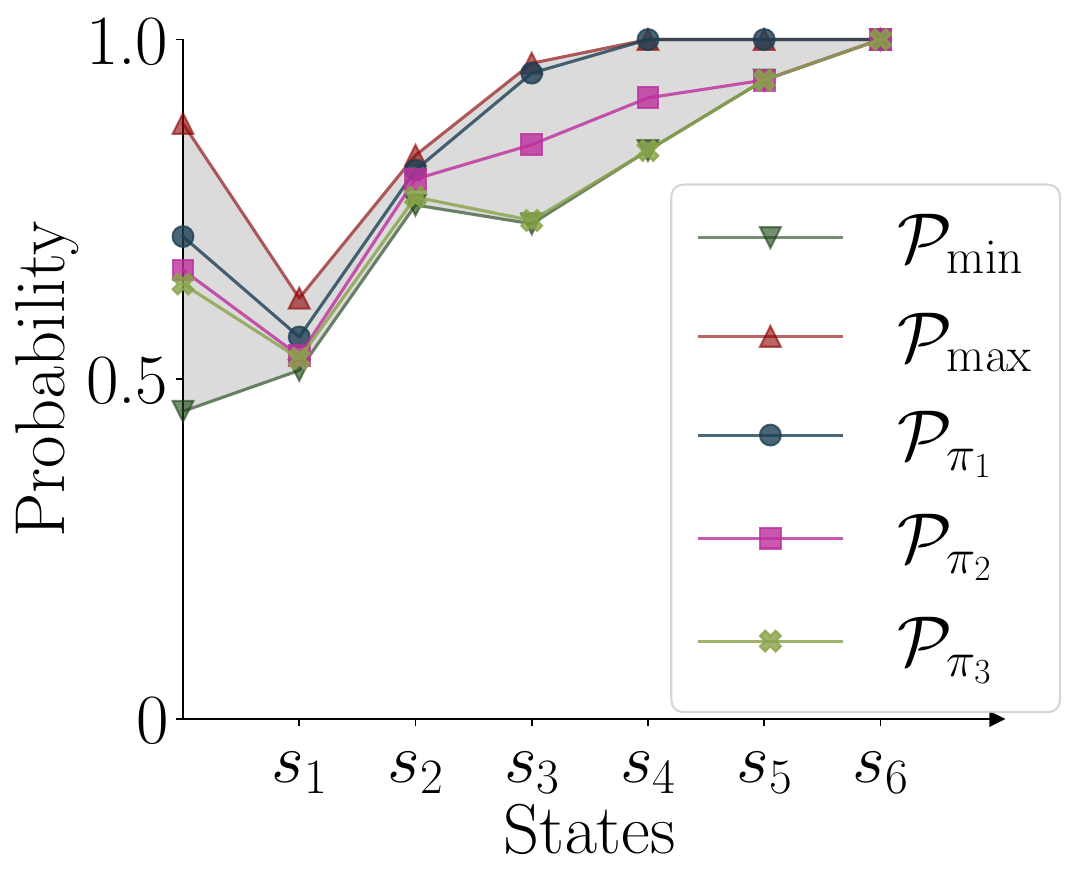}
     \end{subfigure}
     \hfill
     \begin{subfigure}[t]{0.20\textwidth}
         \centering
         \includegraphics[height=3.7cm]{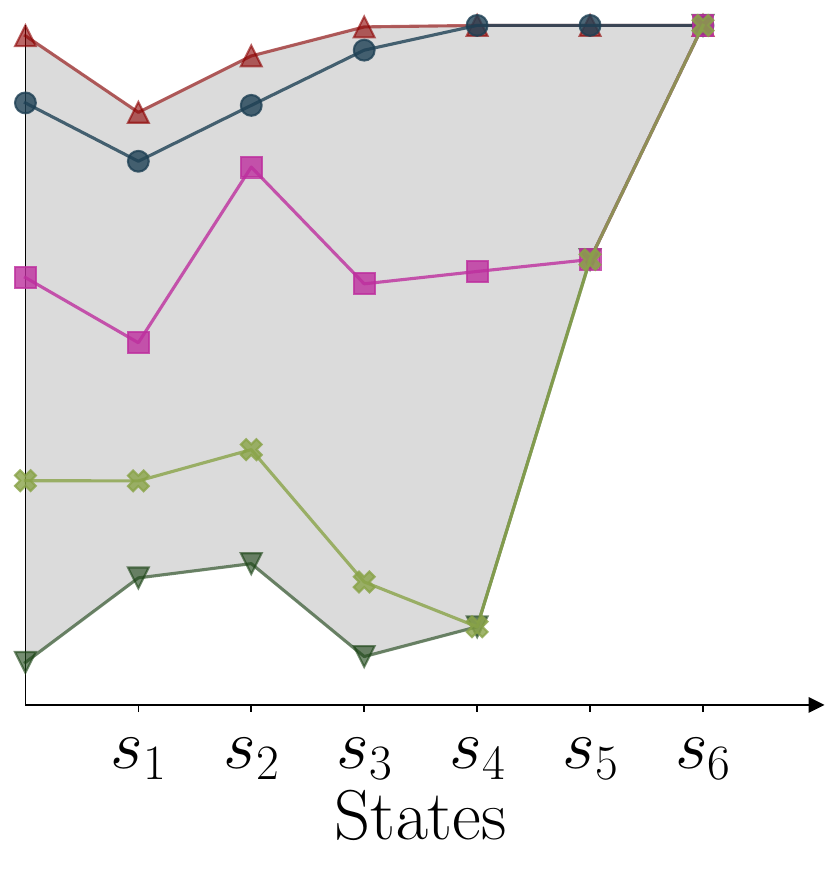}
     \end{subfigure}
        \caption{Comparison of $\tau_{\textit{ref}}$ (left) with a high-agency counterfactual scenario (right).\vspace{1.4em}}
        \label{fig:counterfactualtraces}
\end{figure}

In Figure~\ref{fig:counterfactualtraces} we give the probabilities 
for reaching $S_\mathcal{I}$ for the policies $\pi_1, \pi_2, \pi_3$
for two different traces: left for $\tau_{\textit{ref}}$, right for a counterfactual trace $\tau \in T$ with a high scope of agency.
The figure illustrates how even a single counterfactual trace can be a powerful tool for distinguishing between policies that seem impossible to differentiate with any confidence in the originally given trace $\tau_{\textit{ref}}$.

\begin{table}[b]
\footnotesize
\centering
\begin{tabular}{lrrr}
 & $\pi_1$ & $\pi_2$ & $\pi_3 $  \\
\midrule
$|T|$               & 21    &  100    &  26  \\
$\rho_\pi(T)$       & 0.86  &  0.53    &  0.14  \\
$\sigma_\pi(T)$     & 0.52  &   0.64   &  0.50  \\
\end{tabular}
\caption{Final values of $\rho_\pi(T)$ and $\sigma_\pi(T)$ for different strategies.}
\label{tab:convergence}
\end{table}

A second insight is illustrated in Table~\ref{tab:convergence}.
In this table, for each agent $\pi_1,\pi_2,\pi_3$,
we show the number of counterfactuals needed to generate enough evidence of intentional behavior, 
together with the final values of the intention-quotient and the scope of agency.
Both $\pi_1$ and $\pi_3$ are clear-cut,
but for $\pi_2$ our algorithm reaches the limit of $|T| = 100$ 
without finding enough evidence.
In this case, the intention-quotient of the agent seems to converge to a value of about $0.53$,
sitting in the middle of the lower and upper threshold.

\begin{figure}[t]
    \centering
    \includegraphics[height=3.65cm]{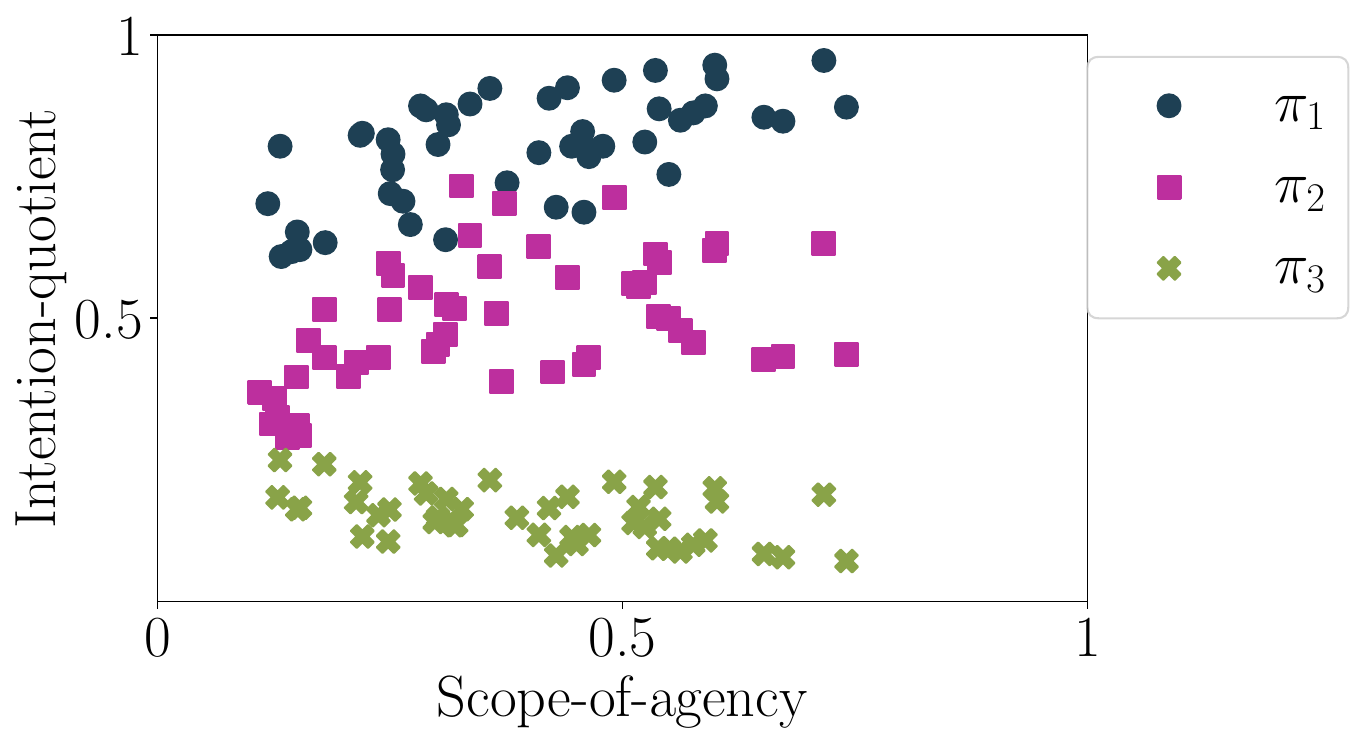}
    \caption{Scatter plot of intention-quotient vs scope of agency for different agents.}
    \label{fig:scatterplot}
\end{figure}

Finally, in Figure~\ref{fig:scatterplot}, 
we show the values of 
intention-quotient against the scope of agency 
for 100 counterfactual traces
sampled from the ranges in Table~\ref{tab:ranges}.
This serves as a visual representation of the same facts presented in Table~\ref{tab:convergence}, concluding that 
$\pi_1$ (\inlinegraphics{figures/p1.png}) is clearly showing evidence of intentionally hitting the pedestrian, 
$\pi_2$ (\inlinegraphics{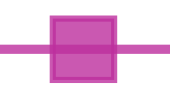}) is showing evidence of intentionally hitting the pedestrian in a lower magnitude, which would be considered enough or not depending on the thresholds,  
and $\pi_3$ (\inlinegraphics{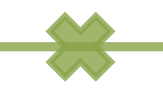}) is showing clear evidence of acting without the intention of hitting the pedestrian.

\section{Discussion}\label{sec:discuss}\

To the best of our knowledge, we present the first method that analyzes intentional behavior directly from the policies given in an MDPs. We believe that our approach has great potential. However, there are aspects that need to be addressed to make the method applicable in challenging scenarios:
\begin{itemize}
    \item Our method requires having a \emph{correct model of the environment} that captures everything relevant to analyze a scenario. In many cases, such models are not available.  
Recent work on digital twin technologies~\cite{jones2020characterising} and the existence of realistic simulators~\cite{DBLP:conf/corl/DosovitskiyRCLK17}
provides optimism for more and more accurate models of agents and their environment.
\item Our method requires the \emph{agent be given as a policy in an MDP}. 
In case we are given a different implementation, \emph{e.g.}, as a neural network, we would need a sample-efficient method to translate the implementation into a policy in the MDP, at least for the relevant parts of the state space.

\item While current probabilistic model checking engines achieve impressive
performance~\cite{budde2021correctness},
\emph{computing exact probabilities is costly} (polynomial complexity).
An alternative would be to use statistical model checking~\cite{10.1145/3158668}, which is less demanding, albeit also less precise.
Statistical model checking has been successfully used to validate autonomous driving modules~\cite{barbier2019validation}.
\end{itemize}

\paragraph{General policies.}
We briefly discuss how to treat policies with memory and non-determinism.
Our definitions naturally extend to non-deterministic policies with memory, 
although it is not evident whether the probabilities required to measure intention-quotients (Definition~\ref{def:intention-quotient}) are easy to compute.
Computing extreme probabilities is equally hard for general policies. 
If the policy has a finite amount $\mu$ of memory, $\mathcal P_{\pi}(\mathtt{Reach}(S_{\mathcal I}),s)$ can be computed using probabilistic model checking, \
with a cost of $\mu$ times that of the memoryless case~\cite{baier2008principles}.
In case the non-determinism is unknown to us, 
to compute $\mathcal P_{\pi}(\mathtt{Reach}(S_{\mathcal I}),s)$
we need to sample the decisions of the agent often enough to get an accurate approximation of its
decision-making probabilities,
making it more costly,
although recent heuristics for determinization may help~\cite{dtcontrol2020}.

\paragraph{Knowledge of the agent's beliefs.}
An intrinsic limitation of studying policies in MDPs 
is the lack of knowledge of the agent's beliefs about the world.
Belief plays a fundamental role in the study of intentions:
an agent that intends $S_{\mathcal I}$ must act believing that their acts 
are a good strategy to reach $S_{\mathcal I}$~\cite{bratman1987intention}.
Belief is also central to the definitions of responsibility and blameworthiness in structural causal models~\cite{chockler2004responsibility,halpern2018towards}.
In part for this reason, together with the uncertainties derived from a probabilistic setting, 
we can only claim incomplete 
evidence of intentional behavior.

\paragraph{Single-agent setting.}
In our framework, all relevant parts of the environment are modeled by an MDP, 
and all the agency in the model is attributed to the agent, 
\ie, the only actor choosing actions in the MDP is the agent.
We argue that this decision is reasonable to study the behavior of an individual agent:
from the perspective of an agent, 
it makes no difference whether the decisions of other actors are governed by a 
sophisticated policy or by random events in the environment, 
as long as the MDP model contains accurate transition probabilities.
The emergence of intrinsically multi-agent phenomena, 
like shared intentions in cooperative settings, would require a multi-agent extension of our framework and is left as future work. 
In particular, we do not explore how to assign moral responsibility to large groups of agents (the so-called ``problem of many hands''~\cite{thompson1980moral,van2015problem}).
Another problem we do not explore is the existence of 
responsibility voids~\cite{braham2011responsibility}, \ie,
situations in which a group of agents should be held accountable for an outcome,
while at the same time, no individual agent intended that outcome.

\section{Related Work}
\paragraph{Intention in artificial intelligence.}
We borrow the concept of intention as a set of states to reach
from standard BDI literature~\cite{rao1991modeling,rao1995bdi}.
Closest related to our work is~\cite{simari2011markov}, 
where the authors develop a mapping from the BDI formalism to the MDP formalism. 
The mapping they propose on intentions to policies in MDPs yields a definition of intentions in MDP similar to our Definition~\ref{def:intentionperfect}
of intentional behavior under perfect knowledge.
In contrast to our work, 
Simari and Parsons focus on optimal policies in MDPs
and their correspondence to plans following a certain intention in the BDI model. 
Therefore, their mapping holds only for optimal policies and cannot be applied to agents with suboptimal policies.
A central element in the definition of intention is
commitment: an agent should not reconsider its intentions too often~\cite{cohen1990intention}.
Although we do not model reconsideration as it relates to time,
the intention quotient $\rho_\pi$ can be interpreted as a quantitative measure of the agent's commitment to reach a certain state.

\paragraph{Responsibility and accountability.}

The concept of intention of rational agents, both humans and non-humans,
has been the subject of extensive study in the context of
philosophy of action~\cite{anscombe2000intention,mele1992springs,bratman1999} as well as in its relation to moral responsibility~\cite{Braham2012,Scanlon2010}.
The concept of agency is a necessary element 
in assigning responsibility, leading to issues when the agency is diluted among many individuals~\cite{shapiro2014massively,braham2011responsibility}.
There is an ongoing debate in the philosophy of mind, between those
that consider that an agent’s reasoning is sufficient to explain
their actions~\cite{Quine1969}, and those who maintain that extrinsic information must be imported through a “Principle of Charity”~\cite{davidson1963actions}. 
By building a model of the agent’s knowledge (the MDP) to inquire about their behavior,
we are assuming the latter position. Recent work attempts to
answer similar questions from the former~\cite{soid}.

\paragraph{Causality and blame attribution.}
A basic element for a complete accountability process is the study of \emph{causality}~\cite{halpern2005causes1,halpern2005causes2}.
The foundational work of~\cite{chockler2004responsibility}
introduced a quantitative notion of causality, 
by studying degrees of responsibility and blame. 
Responsibility and blame allocation has been extensively developed in the context of non-probabilistic structures
(see,  e.g.,~\cite{aleksandrowicz2017computational} for the characterization of complexity or \cite{yazdanpanah2016distant} for a multi-agent framework).
More recent and more closely related to our approach is the work of~\cite{baier2021responsibility}, 
studying responsibility and blame in Markov models.
The study of harm from a causality perspective is also 
gaining attention recently, 
with
\cite{beckers2022causal} studying harm from an actual causality perspective, 
and~\cite{richens2022counterfactual} studying harm from a probabilistic perspective, heavily relying on counterfactuals. 
Counterfactual analysis~\cite{lewis2013counterfactuals} is a key concept in causality~\cite{pearl2009causality},
used in an analogous way as our generation of counterfactual scenarios.
We go one step further by relating the implementation of the agent to the best and worst implementation for reaching an intended event. 
Another recent approach to blame attribution is~\cite{triantafyllou2021blame}, 
which studies multi-agent Markov decision processes from a game-theoretic perspective.

\paragraph{Policy-discovery methods.}
Since the popularization of reinforcement learning,
there exist several methods for obtaining representations of a black-box agent,
by studying traces of such agents. 
In inverse reinforcement learning~\cite{NgR00,10.1145/3158668},
the agent is assumed to be maximizing an unknown reward function, and the objective is to find the reward function that best explains the agent's performance over a set of traces.
These methods could potentially be used as a 
pre-processing step to apply our framework to black box agents.
In any case, the obtained representations must be accurate enough before using them for any accountability process.

\paragraph{Explainability.}
One of the most influential works in \emph{explainability} of AI is~\cite{miller2019explanation},
which studies how explainability should rely on 
concepts from social sciences.
More recently
\cite{winikoff2021bad} 
uses the built-in notions of desire, beliefs and intentions to 
study
explainability of BDI models, relying on concepts from the sociology literature.
While the main paradigm in explainable reinforcement learning is applying techniques from explainable machine learning~\cite{puiutta2020explainable},
our analysis of intentional behavior
can be used as a method to 
aid the interpretability of agents operating in MDPs, using concepts from the philosophy of action~\cite{bratman1987intention}.

\section{Conclusion \& Future Work}
In this paper, we analyzed policies in MDPs with respect to
intentional behavior taking uncertainties into account. Our method uses probabilistic model checking to automatically compute the best and worst possible policy for reaching a set of intended states. 
We assess evidence of intentional behavior in a policy by relating it to the best and worst policies,
and use counterfactual analysis to generate more evidence if needed.

In future work, we want to extend our current analysis by considering a multitude of possibly conflicting intentions of the agent.
Another interesting line of work is to extend the study of intentional behavior to multi-agent systems, in which cooperative or competitive intentions may arise.
We also want to study long executions, where the agent has time for reconsideration.
Furthermore, we want to implement our framework to study reinforcement learning agents in challenging application areas.

\section*{Acknowledgements}
This work was supported in part from the European Union’s Horizon 2020 research
and innovation programme under grant agreement N$^\circ$ 956123 - FOCETA, by the State Government of Styria, Austria – Department Zukunftsfonds Steiermark, the Office of Naval Research (ONR) of the United States Department of Defense through an National Defense Science and Engineering Graduate (NDSEG) Fellowship, and by the National Science Foundation (NSF) awards CCF-2131476 , CCF-2106845, and CCF-2219995.
We also thank Lukas Posch for his help in setting up \textsc{Tempest}.

\bibliographystyle{named}
\bibliography{ijcai23}


\end{document}

%% file: macros.tex
\newcommand{\policy}{\pi}

\newcommand{\ie}{\textit{i.e.}}